\documentclass[10pt,journal,compsoc]{IEEEtran}
\usepackage{rotating}

\usepackage{graphicx}
\usepackage{amsmath,amssymb} 
\usepackage[usenames, dvipsnames]{color}
\usepackage{xcolor}
\usepackage{colortbl}
\definecolor{mygray}{gray}{0.6}
\definecolor{orangepeel}{rgb}{1.0, 0.62, 0.0}
\definecolor{celadon}{rgb}{0.67, 0.88, 0.69}
\definecolor{LightCyan}{rgb}{0.67, 0.88, 0.69}
\definecolor{coralpink}{rgb}{0.97, 0.51, 0.47}
\usepackage{relsize}
\usepackage{multirow,array}
\usepackage[square,sort,comma,numbers]{natbib}

\usepackage{xspace}

\newcommand{\Loss}{\mathcal{L}\xspace}
\newcommand{\nump}{\mathbb{N}^P\xspace}
\newcommand{\nume}{\mathbb{N}^E\xspace}

\newcommand{\edata}{{Aff-Wild}\xspace}

\newcommand{\Ymat}{\ensuremath{\mathbf{Y}}\xspace}

\newcommand{\relationship}{{emotion-to-apparent-personality relationship}\xspace}

\newcommand{\consensus}{\mathcal{G}\xspace}
\newcommand{\domain}{\mathcal{D}\xspace}

\newcommand{\function}{\mathcal{F}\xspace}

\newcommand{\marg}{\ensuremath{m}\xspace}

\newcommand{\proposed}{\emph{PersEmoN}\xspace}

\usepackage{slashbox}
\usepackage{pifont}
\newcommand{\cmark}{\ding{51}}%
\newcommand{\xmark}{\ding{55}}%
\newcommand{\tabincell}[2]{\begin{tabular}{@{}#1@{}}#2\end{tabular}} 
\usepackage[pdftex]{thumbpdf}
\usepackage[switch]{lineno} 
\usepackage{lipsum} 
\usepackage[normalem]{ulem}
  
\usepackage{hyperref}
\hypersetup{
  bookmarks=true,
  bookmarksnumbered=true,
  bookmarksopen=true,
  hypertexnames=false,
  breaklinks=true,
  colorlinks=true,
  linkcolor=red,
  citecolor=blue,
  pagecolor=blue,
  pagebackref=true,
  urlcolor=red
}
\def\zl{}
\def\ZL{}

\begin{document}
%
\title{PersEmoN: \\A Deep Network for Joint Analysis of Apparent Personality, Emotion and Their Relationship}

\author{Le Zhang,
       Songyou Peng$^*$,
        and~Stefan~Winkler,~\IEEEmembership{Fellow,~IEEE}
\IEEEcompsocitemizethanks{
\IEEEcompsocthanksitem Songyou Peng is the corresponding author.\protect
\IEEEcompsocthanksitem Le Zhang is with the Institute for Infocomm Research, Agency for Science, Technology and Research (A*STAR), Singapore.\protect\\
E-mail: zhangl@i2r.a-star.edu.sg
\IEEEcompsocthanksitem Songyou Peng is with the Department of Computer Science, ETH Zurich, and Max Planck ETH Center for Learning Systems. \protect\\
 E-mail: songyou.peng@inf.ethz.ch
\IEEEcompsocthanksitem Stefan Winkler is with the National University of Singapore (NUS).\protect\\
 E-mail: winkler@comp.nus.edu.sg
 \IEEEcompsocthanksitem All the authors are adjunct at the Advanced Digital Sciences Center (ADSC), University of Illinois at Urbana-Champaign, Singapore, where this work was carried out.}}

%
%

\markboth{IEEE Transactions on Affective Computing}%
{}
%




\IEEEtitleabstractindextext{%
\begin{abstract}
Apparent personality and emotion analysis are both central to affective computing. Existing works solve them individually. In this paper we investigate if such high-level affect traits and their relationship can be jointly learned from face images in the wild. To this end, we introduce \emph{PersEmoN}, an end-to-end trainable and deep Siamese-like network. It consists of two convolutional network branches, one for emotion and the other for apparent personality. Both networks share their bottom feature extraction module and are optimized within a multi-task learning framework. Emotion and personality networks are dedicated to their own annotated dataset. Furthermore, an adversarial-like loss function is employed to promote representation coherence among heterogeneous dataset sources. Based on this, we also explore the emotion-to-apparent-personality relationship. Extensive experiments  demonstrate the effectiveness of \emph{PersEmoN}.
\end{abstract}

\begin{IEEEkeywords}
Affective Computing, Emotion, Apparent Personality, Adversarial Learning, Multi-Task Learning, Deep Learning.
\end{IEEEkeywords}}
\maketitle
\IEEEpeerreviewmaketitle

\ifCLASSOPTIONcompsoc
\IEEEraisesectionheading{\section{Introduction}\label{sec:introduction}}
\else
\section{Introduction}
\label{sec:introduction}
\fi

\quad \IEEEPARstart{P}{roliferation} of cameras, availability of cheap storage and rapid developments in high-performance computing have enabled exciting new developments in Human-Computer Interaction (HCI), in which affective computing plays an inevitable role. For instance, in video-based interviews, automatically computed personalities of candidates can serve as an important cue to assess their qualifications. However, affective computing remains a challenging problem in both computer vision and psychology despite many years of research.

\ZL{Facial appearances strongly influence our judgement of the emotion and personality of other people. Such a judgement usually can be made after a very short time~\cite{willis2006first}, although different studies have not yet reached a consensus about the accuracy of such appearance-based first impressions~\cite{naumann2009personality,olivola2010fooled}. As mentioned in a recent survey on personality computing~\cite{vinciarelli2014survey}, state-of-the-art studies consider either the actual personality traits measured from self or acquaintance reports, or the so-called apparent personality traits, which represent the impressions about someone's personality from an external observer's point of view.}

In this paper, we are interested in the problem of analyzing apparent personality, emotion and their relationship. Apparent personality reflects the coherent patterning of behavior, cognition and desires (goals) over time and space, as perceived by an external observer. 
Emotion is an integration of feeling, action, appraisal, and wants at a particular time and location~\cite{revelle2009personality}. 
We can understand the~\relationship as 
weather to climate, i.e.\ what one expects is apparent personality while what one observes in a particular moment is emotion. 
Although they have distinct definitions, the relationship between personality and emotion has been revealed previously.
Eysenck's personality model~\cite{eysenck1950dimensions} showed that \zl{neurotics can be more sensitive to external stimulation and easily become upset or nervous due to minor stressors.  Extraversion on the other hand has been linked to higher sensitivity to potentially rewarding stimuli, which in part explains the high levels of positive affect found in extraverts, since they will more intensely feel the excitement of a potential reward~\cite{depue1999neurobiology}.}


\zl{Apparent personality estimation has become an increasingly popular field of research; related challenges, e.g.\ ChaLearn 2016 on first impressions~\cite{ponce2016chalearn}, together with publicly available datasets, have attracted wide attention. In this paper, we also focus on the task of apparent personality prediction, where we consider the Big Five personality traits} (\emph{Extraversion,
Agreeableness, Conscientiousness, Neuroticism
and Openness})~\citep{ponce2016chalearn}. Our emotion analysis is based on Russell's circumplex model of affect \citep{russell1980circumplex}, in which emotions are distributed in a two-dimensional circular space spanned by the dimensions of \emph{arousal} and \emph{valence}, instead of classifying pre-defined emotion categories.  This is advantageous in the sense that it allows for a finer-grained representation of expressions and emotional states \cite{peng2018deep}.

Deep convolutional neural networks (CNNs) reign undisputed as the new de-facto method for face based applications such as face recognition~\citep{sun2014deep,parkhi2015deep}, alignment~\citep{zhang2016joint}, and so on. This motivates us to study the following fundamental problems:
\begin{enumerate} 
\item As both face recognition and affective computing can have faces as input, how transferable are deeply learned face representations for emotion and apparent personality analysis?
\item Is it beneficial to explore emotion, apparent personality and their relationship in a single deep CNN?  
\end{enumerate}

These tasks are non-trivial. Among the most significant challenges are: 
\begin{itemize}
\item The scarceness of large-scale datasets which encompass both emotion and apparent personality annotations for learning such a rich representation for apparent personality, emotion and emotion-to-apparent-personality relationship. In particular, existing datasets only contain emotion attributes, while other datasets may only be annotated with apparent personality labels. Manually annotating data for both emotion and apparent personality may partly alleviate this. However, it is costly, time-consuming, and error-prone due to subjectivity.

\item The discrepancy of existing datasets: Datasets are usually collected in different  environments  which  may  exhibit  significant  variations  in  illumination, scale, pose, etc. Each dataset may have vastly different statistical distributions.

\item \ZL{Annotations of emotion and/or apparent personality can be done at the image/frame level \cite{guntuku2015others, dhall2016first} or at the video level \cite{ponce2016chalearn}.  How can we encapsulate both frame and video level understanding into a single network? }
\end{itemize}

We address these challenges by proposing~\proposed, an end-to-end trainable and deep Siamese-like network~\citep{bromley1994signature}. It consists of two CNN branches, which we call emotion network and personality network, respectively. Emotion network and personality network share their bottom feature extraction module and are optimized within a multi-task learning framework.  An adversarial-like loss function is further employed to promote representation coherence between heterogeneous dataset sources. We show that ~\proposed works well for analyzing apparent personality, emotion and their relationship. 
A demo version of this paper has been presented in ~\cite{peng2018acmmm}.

\section{Related Work}
The wealth of research in this area is such that we cannot give an exhaustive review.  Instead, we focus on describing the most important threads of research on using deep learning for emotion and apparent personality analysis.

\subsection{Deep Learning for Emotion Analysis}

Emotion analysis has been investigated from different perspectives.  \cite{kim2013deep} proposed a deep belief network for unsupervised audio-visual emotion recognition. However, its feasibility of large-scale supervised learning remains unclear. \cite{surace2017emotion} investigated the usage of deep CNNs and Bayesian classifiers for group emotion recognition in the wild. \zl{Apart from visual inputs, the system also needs the scene context (such as background, clothes, etc.) which may not be available in many scenarios.}  \cite{ranganathan2016multimodal} introduced convolutional deep belief networks to learn salient multi-modal features of emotions. \zl{Although workable as reported, their network structure is shallow, and it remains unclear how to transfer rich feature hierarchies from very deep networks for different modalities in their system.}

Unlike popular classification approaches for discrete emotion categories, many recent works delve into different representations of human expressions and emotions. For instance, EmotioNet~\citep{fabian2016EmotioNet} provides an accurate, real-time algorithm for the automatic annotation of a million facial expressions in the wild. \zl{However, the performance gap between using EmotioNet and human-annotated datasets for training emotion networks is not well understood.} Other approaches try to analyze emotion using continuous arousal-valence  space~\citep{russell1980circumplex}. For instance, Mollahosseini \emph{et al.}~\cite{mollahosseini2017affectnet} introduce a large-scale emotion dataset and show that their deep neural network outperforms conventional machine learning methods and off-the-shelf facial expression recognition systems. \zl{However, it requires a specially designed sampling strategy to alleviate data imbalance problems}.
\ZL{An ensemble of memory networks~\citep{li2017estimation}, multiple datasets for cascade learning~\citep{chang2017fatauva}, and multiple LSTM layers~\citep{hasani2017facial} have been employed to predict the emotion scores. Due to the complexity of these networks, they are difficult to train however.}


\subsection{Deep Learning for Apparent Personality Analysis}    

\ZL{G{\"u}{\c{c}}l{\"u}t{\"u}rk \emph{et al.} \cite{guccluturk2016deep} introduced a deep audio-visual residual network for multimodal apparent personality trait recognition. In their extended version \cite{guccluturk2017multimodal}, the authors analysed different cues (visual, acoustic and audiovisual). This may not be very practical as some of the cues could be missing during deployment of the system.  \cite{subramaniam2016bi} developed a volumetric convolution and Long-Short-Term-Memory (LSTM) based network to learn audio-visual temporal patterns. Although it outperforms many other approaches~\cite{ponce2016chalearn}, its performance on pure visual inputs is not reported, which makes it difficult to understand the merits of their visual stream.} 

\cite{zhang2016deep} identified apparent personality with a deep bimodal regression framework based on both video and audio inputs. \zl{However, the performance of all above-mentioned methods relies heavily on ensemble strategies, whereas we are able to achieve better results with a single visual stream with the proposed~\proposed.} 

G{\"u}rp{\i}nar~\emph{et al.}~\cite{gurpinar2016combining} employed a pre-trained CNN to extract facial expressions as well as ambient information for apparent personality analysis. \zl{Although they achieve promising results, the system is not end-to-end trainable and needs a stand-alone regressor.} \zl{A time-continuous
prediction approach that learns the temporal relationships, rather than treating each time instant separately, was  established in~\cite{celiktutan2017automatic} for the prediction of Big Five traits, Attractiveness and Likeability. Nevertheless, a database annotated in a time-continuous manner is needed in their setting, which is difficult to obtain in practice. }

\zl{Ventura~\emph{et al.}~\cite{ventura2017interpreting} investigated the reason of the success of CNN in apparent personality analysis. They showed that the face provides most of the discriminative information for this task. This motivated us to investigate the feasibility of using a very deep face verification network for this task.} For more related work on apparent personality analysis, please refer to recent surveys~\citep{junior2018first,escalante2018explaining}.

\section{Methodology}
In comparison to the aforementioned studies, our work aims to investigate whether emotion and apparent personality analysis can benefit from the face representations learned from a well-annotated face recognition dataset, \emph{without having a dataset with both emotion and apparent personality annotations}. To this end, we show that state-of-the-art face recognition networks perform well for both emotion and apparent personality analysis. We also explore the feasibility of jointly training emotion and apparent personality analysis. More specifically, we propose \proposed within a multi-task learning framework to learn better representations for both emotion and apparent personality than those obtained by solving each task individually. On top of such representations, we demonstrate the feasibility of establishing a good~\relationship. 

\subsection{\proposed Overview}
\label{sec:overview}
\begin{figure*}
	\centering
    \includegraphics[width=0.8\textwidth]{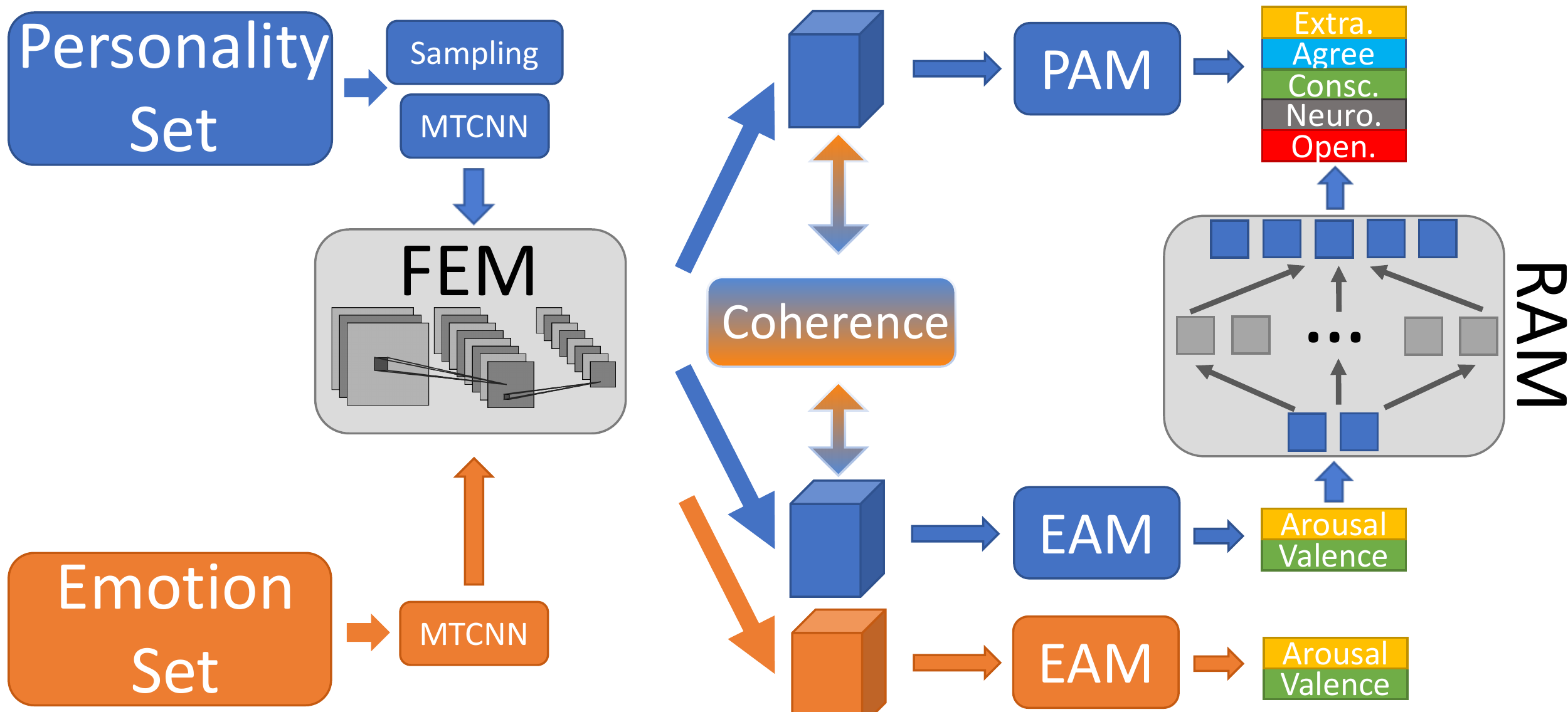}
    \caption{Workflow of the proposed \proposed. \ZL{The colours in the diagram represent the data flow of the system. In the training phase, the system is aware of which dataset the image comes from and will automatically assign the image to its own branch. During inference, we can use \emph{FEM}  and \emph{PAM} to obtain the apparent personality traits. Similarly, we can use \emph{FEM}  and \emph{RAM} to obtain the emotion outputs. The boxes preceding \emph{PAM} and \emph{EAM} are the feature representations. Please refer to the text for more details.}}
    \label{fig:overview}
\end{figure*}

An overview of \proposed can be found in Fig.~\ref{fig:overview}. We first detect and align faces for both apparent personality and emotion datasets with an open version of \emph{MTCNN}~\citep{zhang2016joint}. For the apparent personality dataset, we employ a sparse sampling strategy. The personality network consists of a feature extraction module (\emph{FEM}) and a personality analysis module (\emph{PAM}) to predict the Big Five personality factors.  A consensus aggregation function is employed to aggregate raw apparent personality scores before feeding them into the \emph{PAM}. Similarly, the emotion network shares the \emph{FEM} with the personality network and has its own emotion analysis module (\emph{EAM}) targeted at predicting the arousal and valence dimensions~\citep{zafeiriou2017aff} of emotion. An \relationship analysis module (\emph{RAM}) is also employed. 

\zl{In the training phase, the system is aware of which dataset the image comes from and will automatically assign the image to its own branch.}  \ZL{For instance, the images from the apparent personality set go through \emph{FEM}  and PAM. Meanwhile, they can also go through  \emph{FEM}  and \emph{EAM} and finally output the apparent personality traits through  \emph{RAM}. In the same way, the images from the emotion set can go through \emph{FEM}  and \emph{EAM} to yield emotion outputs.}   

\ZL{In the testing phase, the system can estimate the emotion and the apparent personality from \emph{EAM} and \emph{PAM} separately. During inference, we could use \emph{FEM}  and \emph{PAM} to obtain the apparent personality traits. Similarly, we could use  \emph{FEM}  and  \emph{RAM} to get the emotion outputs. As a side product, we could even use  \emph{RAM} to produce the apparent personality attributes from emotion (arousal and valence) inputs.  Note that in the testing phase the proposed method also works with video-based emotion datasets by processing each video frame individually.} 

The detailed network structure of the various modules of~\proposed is summarized in Table~\ref{tab:sphereface}. In the following section, we elaborate on the different components mentioned above.

\begin{table}[!thb]
\centering
\caption{ \zl{Detailed architecture of \proposed. Conv denotes convolution units that may contain multiple convolution layers; residual units are shown in square brackets. For example, $[3×3, 64]\times 4$ denotes 4 cascaded convolution layers with 64 filters of size 3×3, and $S2$ denotes stride 2. FC is a fully connected layer, for which the number of output neurons are reported.}}
\begin{tabular}{c|c|c}\hline
 Module&Layer &  Details    \\\hline
&Conv 1&\tabincell{c}{$[3\times 3,64]\times1 ,S2$\\ $\begin{bmatrix} 3\times3,64\\  3\times 3,64 \end{bmatrix} \times 1$}\\ \cline{2-3}
& Conv 2&\tabincell{c}{$[3\times 3,128]\times1 ,S2$\\ $\begin{bmatrix} 3\times3,128\\  3\times 3,128 \end{bmatrix} \times 2$}\\ \cline{2-3}
&  Conv 3&\tabincell{c}{$[3\times 3,256]\times1 ,S2$\\ $\begin{bmatrix} 3\times3,256\\  3\times 3,256 \end{bmatrix} \times 4$}\\ \cline{2-3}
& Conv 4&\tabincell{c}{$[3\times 3,512]\times1 ,S2$\\ $\begin{bmatrix} 3\times3,512\\  3\times 3,512 \end{bmatrix} \times 1$}\\ \cline{2-3}
\multirow{-5}{*}[3.5em]{\textbf{{FEM}}} &  FC1&512\\ \hline
& FC2& 5\\\cline{2-3}
\multirow{-2}{*}{\textbf{{PAM}}}& Pooling & AVE\\\hline
\textbf{EAM}& FC3& 2\\\hline
& FC4 &128\\\cline{2-3}
\multirow{-2}{*}{\textbf{{RAM}}}& FC5 & 2\\\hline
\textbf{Coherence} &FC6 &2\\\hline
 \end{tabular}
 \label{tab:sphereface}
\end{table}
\subsection{Personality and Emotion Networks}
\label{sec:PEN}
A shared \emph{FEM}, embodied with a truncated~\emph{SphereFace} network~\citep{liu2017sphereface} with its last two layers removed, is employed for both branches. Those two branches are dedicated to emotion and apparent personality-annotated datasets, respectively, and jointly optimized with the \emph{FEM}. 

To utilize rich information from each video frame for more effective network training, the personality network operates on a pool of sparsely sampled face frames from the entire video. Each face frame in this pool can produce its own preliminary prediction of the apparent personality score. We take inspiration from recent advances in video based human action recognition~\citep{wang2016temporal} and employ a consensus strategy among all the face frames from each video to give a video-level prediction on the apparent personality. The loss values of video-level predictions, other than those of face-level ones, are optimized by iteratively updating the model parameters. We use $V$ and $\Ymat$ to represent a generic video input and its ground truth label. Given the $i^{th}$ video $\{V^P_i,\Ymat^P_i\}(i\in \nump)$, where $\nump$ stands for the index set of apparent personality videos, and $P$ denotes the data source, i.e.\ apparent personality dataset here, we divide it into $K$ segments $\{S^P_{i1}, S^P_{i2}, \cdots , S^P_{iK}\}$ of equal duration. Now our personality network models a sequence of faces as follows:
\begin{equation}
\begin{aligned}
\mathbf{P}(V_i,W^P)=&\mathbf{P}(I^P_{i1},I^P_{i2},\cdots,I^P_{iK},W^P)\\
     =&\consensus(\function(I^P_{i1},W^{P}),\function(I^P_{i2},W^{P}),\cdots,\function(I^P_{iK},W^{P}))
\end{aligned}
\label{equ:pn}
\end{equation}
Here $(I^P_{i1}, I^P_{i2},\cdots, I^P_{iK})$ is a pool of face frames, where each face $I^P_{ik}$ is randomly
sampled from its corresponding segment $S^P_{ik}$. The function $\function(I^P_{ik},W^{P})$ represents the
personality network with parameters $W^{P}$ which operates on face $I^P_{ik}$ and provides preliminary apparent personality scores. The segmental consensus function $\consensus$ aggregates the raw outputs from multiple face frames to obtain a final apparent personality score for each video. Although the proposed method is generic and applicable for a wide range of functions such as \emph{max}, \emph{average}, \emph{recurrent aggregation}, we use the \emph{average} function similar to~\cite{wang2016temporal}.  Based on this consensus, we optimize the personality network with the smooth $\ell_1$ loss function~\citep{girshick2015fast} defined as:
\begin{equation}
\Loss_{per}(W^P)= \sum\limits_{i\in \nump} \text{smooth}_{\ell_{1}}(\Ymat^P_i-\mathbf{P}(V_i^P,W^P))
\end{equation}
The smooth $\ell_1$ function is given below; $m$ represents a margin parameter.
\begin{equation}
\text{smooth}_{\ell_{1}}(x)=
\begin{cases}
 \frac{1}{2}(x)^2 & |x|<\marg,\\ 
 |x|-0.5&  \textrm{otherwise.}
\end{cases}
\label{eq:smooth}
\end{equation}

The emotion network works in a simpler manner by directly processing input faces, since frame level annotations are already available. More specifically, given a face image $\{I^E_i,\Ymat^E_i\} (i\in\nume)$, the emotion network produces emotion scores as:
\begin{equation}
\begin{aligned}
\mathbf{E}(I^E_i, W^E)=\function(I^E_i,W^{E})
\end{aligned}
\label{equ:en}
\end{equation}
Similarly, the loss function for the emotion network is:
\begin{equation}
\Loss_{emo}(W^E)=\sum\limits_{i\in \nume} \text{smooth}_{\ell_{1}}(\Ymat^E_i-\mathbf{E}(I^E_i,W^E))
\label{equ:enet}
\end{equation}

\subsection{Representation Coherence}

People may appear in various scales and poses under different illumination conditions \ZL{for different datasets}. Besides, each dataset may exhibit different statistical distributions \ZL{ and annotation bias}. Representations learned from each \ZL{dataset} individually without pursuing coherence between them may present significant discrepancy. A representation with good transferability should be \ZL{dataset}-invariant in the sense that the learned representations are coherent for different data samples from different \ZL{datasets}~\citep{tzeng2015simultaneous}.  This is also beneficial to exploring the~\relationship in our case. To this end, a classifier is trained to classify which dataset the input image comes from. After convergence of the system, it cannot distinguish them because the final representation is dataset-invariant. \ZL{This strategy reduces over-fitting by learning a generalizable representation, which is applicable not only to the
tasks in question, but also to other tasks with significant commonalities. In our setting, since a shared network backbone is employed by two tasks, additional tasks act as a regularization which requires the system to perform
well on a related task.}

We take inspiration from~\cite{tzeng2015simultaneous} by training a dataset classifier, denoted as $\domain$ with parameters $W^{\domain}$, to perform binary classification to distinguish which dataset a particular datum comes from. For each feature representation from the \emph{FEM}, we learn the dataset classifier with the following softmax loss. \zl{In each mini-batch, the loss is as follows when sampling from the personality dataset:
\begin{equation}
    \Loss_\domain^{P}(W^{\domain}) = -\sum\limits_{i\in\nump}\sum\limits_{k=1}^{K}\log q(I^P_{ik}, W^P,W^\domain)
\end{equation}
where $q(I, W,W^\domain) = \mathrm{softmax}(W^{\domain}\boldsymbol{\cdot} \function(I,W))$. Similarly, the loss for samples $j$ from the emotion dataset is:
\begin{equation}
    \Loss_\domain^{E}(W^{\domain}) = -\sum\limits_{j\in\nume}\log q(I^E_{j}, W^E, W^\domain)
\end{equation}
The overall loss for each mini-batch can be computed:
\begin{equation}
    \Loss_\domain(W^{\domain}) = \Loss_\domain^{P}(W^{\domain})+\Loss_\domain^{E}(W^{\domain})
\end{equation}}


As in \citep{tzeng2015simultaneous}, an adversarial-like learning objective is introduced in  the \emph{FEM} which aims at ``maximally confusing'' the two datasets by computing the cross entropy between the output predicted dataset labels and a uniform distribution over dataset labels:
\begin{equation}
\begin{split}
\Loss_{adv}(W^P,W^E) = -&\sum\limits_{i\in \nump}\sum\limits_{k=1}^{K}\log q(I^P_{ik}, W^P,W^\domain)\\ 
-&\sum\limits_{i\in \nume}\;\;\; \log q(I^E_i, W^E, W^\domain)\\
+& \log q(I^P_{ik}, W^E,W^\domain)\\
+&\log q(I^E_i, W^P, W^\domain)
\end{split}
\label{equ:adver}
\end{equation}
Similar to the adversarial-learning, we perform iterative updates for both $\Loss_\domain(W^{\domain})$ and $\Loss_{adv}(W^P,W^E)$ given the fixed parameters from the previous iteration. 

\subsection{Emotion-to-Apparent-Personality Relationship Analysis}
Here we investigate whether apparent personality can be inferred directly from emotion attributes. This is challenging due to the paucity of datasets which encompass both emotion and apparent personality annotations for us to learn such a relationship. We insert a relationship analysis module (\emph{RAM}), which receives the emotion scores from the emotion analysis network and predicts apparent personality scores. More specifically, the input of~\emph{RAM} can be obtained by:
\begin{equation}
\begin{aligned}
\mathbf{E}(V_i^P,W^E)=&\mathbf{E}_i=\mathbf{E}(I_{i1}^P,I_{i2}^P,\cdots,I_{iK}^P,W^E)\\
     =&(\function(I_{i1}^P,W^{E}),\function(I_{i2}^P,W^{E}),\cdots,\function^{E}(I_{iK}^P,W^{E}))
\end{aligned}
\end{equation}
As we already defined,  $(I_{i1}^P, I_{i2}^P,\cdots, I_{iK}^P)$ is a pool of faces from the apparent personality dataset where each face $I_{ik}^P$ is randomly sampled from its corresponding segment $S_{ik}^P$. $\function(I_{ik}^P,W^{E})$ represents the
emotion network with parameters $W^{E}$ which operates on face $I_{ik}^P$ to give preliminary results on the emotion scores. \emph{RAM} employs the same consensus strategy among all the faces from the video to output the aggregated apparent personality score $\mathbf{R}$ of video $V_i^P$:
\begin{equation}
\mathbf{R}(\mathbf{E_i},W^\mathbf{R})=\mathbf{R}(\mathcal{G}(\mathbf{E}_i),W^\mathbf{R}),
\end{equation}
where $W^{\mathbf{R}}$ represents the weights of \emph{RAM}. \emph{RAM} is trained by optimizing the following objective function:
 \begin{equation}
\Loss_{RAM}(W^\mathbf{R})=\sum\limits_{i\in \nump}\text{smooth}_{\ell_{1}}(\Ymat^P_i-\mathbf{R}(\mathbf{E}_i,W^\mathbf{R}))
\label{equ:ramnet}
\end{equation}


\subsection{Overall Loss Functions}
Every module of \proposed is differentiable, allowing end-to-end optimization of the whole system.  The learning process of \proposed aims to minimize the following loss:
\begin{equation}
\begin{split}
\Loss=\lambda_1&\Loss_{per}(W^P)+\lambda_2\Loss_{emo}(W^E)+\lambda_3\Loss_\domain(W^{\domain})\\
&+\lambda_4\Loss_{adv}(W^P,W^E)+\lambda_5\Loss_{RAM}(W^\mathbf{R})
\end{split}
\end{equation}


\section{Experiments}
\subsection{Dataset and Evaluation Protocol}
\begin{figure}
	\centering
    \includegraphics[width=0.4\textwidth]{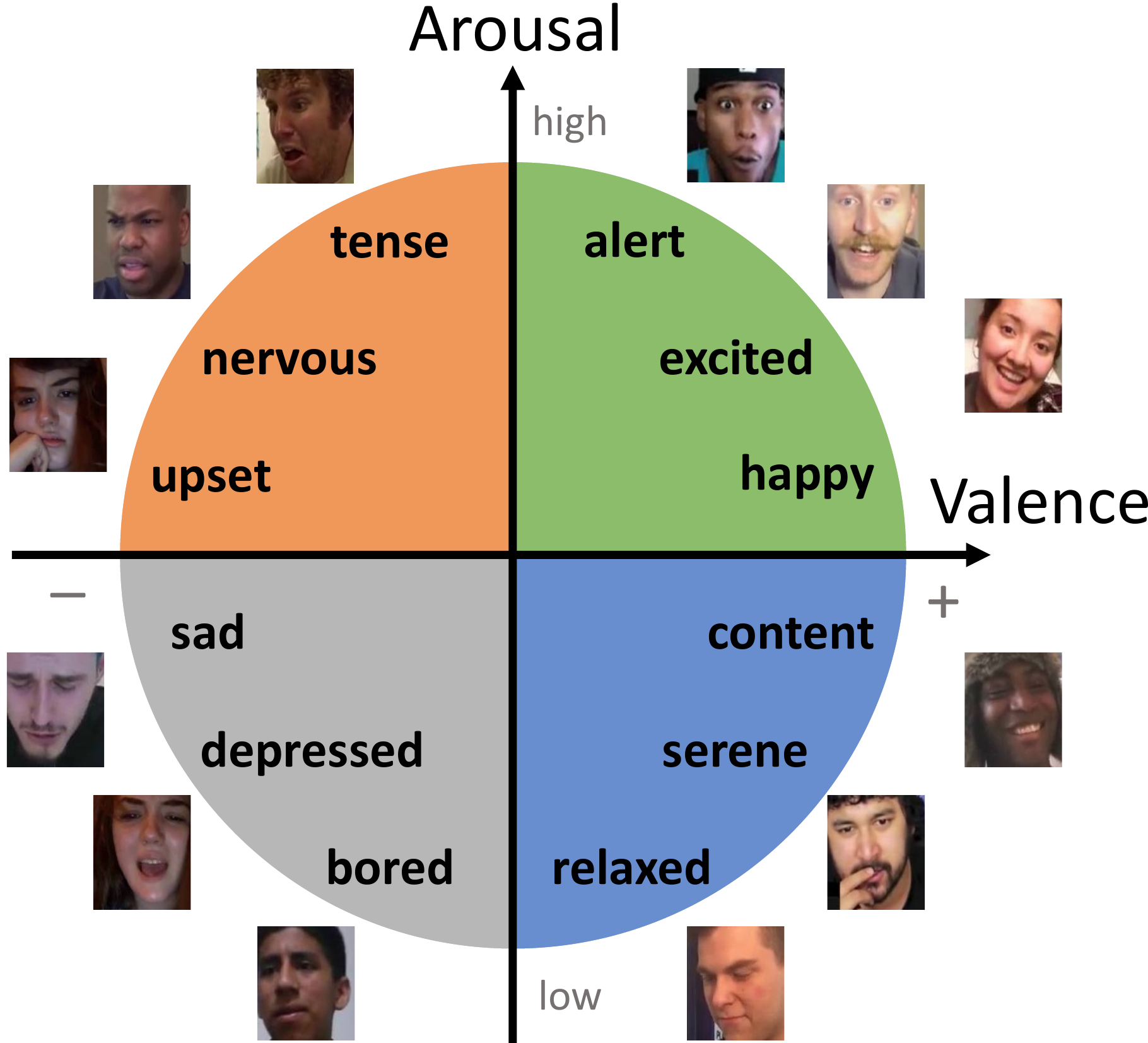}
    \caption{Emotion wheel showing the connection between emotion categories and arousal-valence space.}
    \label{fig:affective-circle}
\end{figure}

We choose two large challenging datasets to evaluate~\proposed. The Aff-Wild emotion dataset \citep{zafeiriou2017aff} consists of 298 YouTube videos (252 for training and 46 for testing) with a total length of about 30 hours (over 1M frames). The videos show the reaction of individuals to various clips from movies, TV series, trailers, etc. Each video is labeled by 6-8 annotators with frame-wise valence and arousal values, with a total of $200$ annotators. Both valence and arousal values range from $-1$ to $1$.
The representation of emotions via arousal/valence values is illustrated in Fig.~\ref{fig:affective-circle}. For apparent personality, we use the ChaLearn personality dataset~\citep{ponce2016chalearn}, which consists of $10k$ short video clips with 41.6 hours (4.5M frames) in total. In this dataset, people face and speak to the camera. Each video is annotated with apparent personality attributes as the Big Five personality traits in $[0,\;1]$.  
The annotation was done via Amazon Mechanical Turk.


To assess the quality of emotion predictions from ~\proposed, \zl{we calculate the mean square error (MSE) between the predicted values of \ZL{emotion scores} and the ground truth.}
For the evaluation of the apparent personality recognition, we apply two metrics used in the ECCV 2016 ChaLearn First Impressions Challenge~\citep{ponce2016chalearn}, namely mean accuracy $A$ and coefficient of determination $R^2$, which are defined as follows:
\begin{equation}
	A = 1 - \frac{1}{N^t} \sum_{i}^{N^t} |\Ymat^P_i - \mathbf{P}_i|, 
    \label{eq:mean_accuracy}
\end{equation}
\begin{equation}
	R^2 = 1 - \sum_{i}^{N^t} (\Ymat^P_i - \mathbf{P}_i)^2/ \sum_{i}^{N^t} (\mathbf{\bar{Y}}^P -\mathbf{P}_i)^2
    \label{eq:Rtwo}
\end{equation}
where $N^t$ denotes the total number of testing samples, $\Ymat^P$ the ground truth, $\mathbf{P}_i$ the prediction, and $\mathbf{\bar{Y}}^P$ the average value of the ground truth.

\subsection{Implementation}
We initialize~\emph{FEM} with a truncated 20 layer version of the \emph{SphereFace} model~\citep{liu2017sphereface}. 
\emph{PAM} is embodied with a fully connected (FC) layer with 5 outputs, while \emph{EAM} has only 2 output neurons in the FC layer. We use $sigmoid$ and $tanh$ to squash the outputs for \emph{PAM} and \emph{EAM} respectively.  We use a single-hidden-layer feed-forward network to analyze the \relationship.  More specifically, \emph{RAM}  is implemented with two FC layers where the first one receives 2 emotion scores as input and output 128 features with ReLU nonlinearity. The same consensus function and $sigmoid$ nonlinearity are used to obtain the apparent personality traits for RAM.  Additional architecture details are provided in Table~\ref{tab:sphereface}.


\proposed is implemented in Caffe~\citep{jia2014caffe}. We train the whole network with an initial learning rate of $0.01$.  For each mini-batch, we randomly select 100 images from the Aff-Wild dataset and 10 videos from Chalearn. For each video, 10 frames are further sparsely sampled in a randomized manner, i.e.\ $K=10$. Hence, the overall batch size is equal to 200. We train the network for $56k$ iterations and decrease the learning rate by a factor of 10 in the $32k^{th}$ and $48k^{th}$ iteration. 
\zl{As the main goal of the system is to estimate emotion and apparent personality attributes, $\mathcal{L}_{emo}$ and $\mathcal{L}_{per}$ should be the main objective functions, and hence their weights are set to $\lambda_1=\lambda_2=1$. Other loss functions serve as regularizers to further improve the results, and hence their weights are set to relatively smaller values, $\lambda_3=\lambda_4=\lambda_5=0.1$.}  The margin parameter in all the smooth $\ell_1$ loss (Eq.~\eqref{eq:smooth}) is set to $\marg=0.05$. 

\subsection{Evaluation of Emotion}
We first report the results of emotion prediction on the \edata dataset. \proposed is compared with a strong baseline method CNN-M and $3$ benchmark methods from the Aff-Wild challenge~\citep{zafeiriou2017aff}. \ZL{ More specifically, MM-Net~\cite{li2017estimation} consists of a carefully designed deep face feature learner to learn discriminative features for affective levels and then employs multiple memory networks for feature aggregation. FATAUVA~\citep{chang2017fatauva} first learns the facial part-based response through attribute recognition CNNs, which is further used to supervise the learning of action unit (AU) detection. Finally, it employs AUs as a  mid-level representation to estimate the intensity of valence and arousal. DRC-Net~\cite{hasani2017facial} is based on Inception-ResNet modules redesigned specifically for the task of facial affect estimation. It consists of a shallow Inception-ResNet, a deep Inception-ResNet and an inceptionResNet with LSTMs. These networks extract facial features at different scales and simultaneously estimate both valence and arousal in each frame.}

\ZL{Since annotations of the test data are not public, our results in Table~\ref{tab:emotion} were evaluated by the official organizer.} A total of 9 evaluations were obtained from the organizers. As demonstrated in Table~\ref{tab:emotion}, our method achieves competitive accuracy to these state-of-the-art methods on the test data.
\begin{table}
\centering
\caption{Results (MSE) of emotion task on \edata.}
\label{tab:emotion}
\begin{tabular}{rcc}\hline
 Method&Arousal&Valence\\\hline
 CNN-M~\citep{zafeiriou2017aff}&0.140&0.130\\
 MM-Net~\citep{li2017estimation}& \textbf{0.088}&0.134\\
 FATAUVA~\citep{chang2017fatauva}&0.095&\textbf{0.123}\\
 DRC-Net~\citep{hasani2017facial}&0.094&0.161\\
 \textbf{\proposed} (ours)&0.108&0.125\\\hline
\end{tabular}
\vspace{-1em}
\end{table}

Simplicity is central to our design; the strategies adopted in~\proposed are complementary to those more complicated approaches, such as ensemble of memory networks used in MM-Net, multiple datasets used for cascade learning employed in FATAUVA-Net and multi-scale inputs adopted in DRC-Net. 
Furthermore, all these other methods are much more difficult to train than ours. Multiple LSTM layers are used in MM-Net and DRC-Net, while FATAUVA-Net cannot perform end-to-end but cascade training. \zl{Although \proposed was not optimized for emotion recognition like the other methods, it still yields competitive results for the emotion task.}

\subsection{Evaluation of Apparent Personality}

Recognition of Big Five personality traits appears more interesting to us because apparent personality is a higher-level attribute compared to emotion. 
Table~\ref{tab:chalearn-properties} compares some of the latest apparent personality recognition methods.
In contrast to other approaches, ours can be trained end-to-end using only one pre-trained model.
Moreover, unlike most methods which fuse both acoustic and visual cues, ~\proposed uses only video as input.

\begin{table}[!htb]
\centering
\renewcommand{\tabcolsep}{3pt}
\caption{Comparison of the deep personality network properties of \proposed vs.\ the top teams in the 2016 ChaLearn First Impressions Challenge.}
\small
\begin{tabular}{rcccc}
\hline
& \multirow{2}*{\textbf{Fusion}} & \multicolumn{2}{|c|}{\textbf{Modality}}  & \multirow{2}*{\textbf{End-to-End}} \\\cline{3-4}
&&\textbf{Audio}&\textbf{Video}& \\\hline
\textbf{\proposed}\;\;  & late &  \xmark   & \cmark & \cmark        \\
NJU-LAMDA~\cite{zhang2016deep}$^1$ & late  &  \cmark   &  \cmark   & \cmark \\
evolgen~\cite{subramaniam2016bi}\;\;   & early   & \cmark & \cmark & \cmark        \\
DCC~\cite{guccluturk2016deep}\;\;       & late    & \cmark & \cmark &  \cmark          \\
ucas~\cite{ponce2016chalearn}\;\;       & late & \cmark & \cmark & \xmark \\
BU-NKU-v1~\cite{gurpinar2016combining}\;\; & early    & \xmark & \cmark & \xmark \\
BU-NKU-v2~\cite{gurpinar2016multimodal}$^2$ & early   & \cmark & \cmark & \xmark \\\hline
\end{tabular}
\vspace{-0.7em}\\

\bigskip {\scriptsize $^1$ winner, $1^{st}$ ChaLearn First Impressions Challenge (ECCV 2016). \\$^2$ winner, $2^{nd}$ ChaLearn First Impressions Challenge (ICPR 2016)}
\label{tab:chalearn-properties}
\end{table}

The quantitative comparison between \proposed and state-of-the-art works on apparent personality recognition is shown in Table~\ref{tab:chalearn-results}. 
The teams from NJU-LAMDA to BU-NKU-v1 are the top five participants in the $1^{st}$ ChaLearn Challenge on First Impressions~\citep{ponce2016chalearn}.
Note that BU-NKU was the only team not using audio in the challenge, and their predictions were rather poor comparatively.
After adding the acoustic cues, the same team won the $2^{nd}$ ChaLearn Challenge on First Impressions~\citep{escalante2016chalearn}.
Importantly, \proposed only considers visual streams. Yet as is evident in Table~\ref{tab:chalearn-results}, even when only taking into account \emph{PAM}, \proposed already achieves superior performance over others, not only on the average $A$ and $R^2$ scores, but both scores for all traits.  

\begin{table*}[ht!]
\centering
\renewcommand{\tabcolsep}{5pt}
\caption{Apparent personality prediction benchmarking using mean accuracy $A$ and coefficient of determination $R^2$. Note that there are no $R^2$ scores reported for BU-NKU-v2.}
\label{tab:personality-comp}
\begin{tabular}{c|cc|cc|cc|cc|cc|cc}
\hline
            {}  & \multicolumn{2}{c|}{Average}       & \multicolumn{2}{c|}{\textit{Extraversion}}   & \multicolumn{2}{c|}{\textit{Agreeableness}} & \multicolumn{2}{c|}{\textit{Conscientiousness}} & \multicolumn{2}{c|}{\textit{Neuroticism}} &  \multicolumn{2}{c}{\textit{Openness}} \\ \hline 
& $A$ &  $R^2$          & $A$ &  $R^2$ & $A$ &  $R^2$           & $A$ &  $R^2$& $A$ &  $R^2$& $A$ &  $R^2$\\ \hline   \hline         
\rowcolor{orangepeel}
\textbf{\emph{PAM}+\emph{RAM}} & \textbf{0.917} & \textbf{0.485} & \textbf{0.920}&\textbf{0.552} & \textbf{0.914}&\textbf{0.349} & \textbf{0.921} & 0.570  & \textbf{0.914} & \textbf{0.500}  & \textbf{0.915} & \textbf{0.457}              \\ \hline
\rowcolor{LightCyan}
Ours (\emph{PAM}) & 0.916 &0.478 &  \textbf{0.920} & 0.544 & 0.913 & 0.338 & \textbf{0.921}&\textbf{0.571}  & 0.913 &0.489    & 0.914 &0.448               \\ \hline
Ours (\emph{RAM}) & 0.903&0.373 & 0.911&0.449& 0.908&0.264 & 0.902 & 0.349 & 0.908 & 0.442 &  0.907 & 0.364 \\ \hline
NJU-LAMDA~\cite{zhang2016deep}     & 0.913 & 0.455          & 0.913 & 0.481          & 0.913 & 0.338        & 0.917 & 0.544            & 0.910 &  0.475                           & 0.912 & 0.437         \\ \hline
evolgen~\cite{subramaniam2016bi}       & 0.912 &  0.440          & 0.915 &  0.515          & 0.912 & 0.329           & 0.912 & 0.488            & 0.910 & 0.455                           & 0.912 & 0.414                        \\ \hline
DCC~\cite{guccluturk2016deep}           & 0.911 & 0.411          & 0.911 & 0.431          & 0.910& 0.296          & 0.914 & 0.478            & 0.909 & 0.448                           & 0.911 & 0.402                        \\ \hline
ucas~\cite{ponce2016chalearn}          & 0.910 & 0.439          & 0.913 & 0.489          & 0.909 & 0.292          & 0.911 & 0.520            & 0.906 & 0.457                           & 0.910 & 0.439                        \\ \hline
BU-NKU-v1~\cite{gurpinar2016combining}     & 0.909 & 0.394          & 0.916 & 0.514          & 0.907 & 0.234          & 0.913 & 0.487            & 0.902 & 0.363                           & 0.908 & 0.372                        \\ \hline
BU-NKU-v2~\cite{gurpinar2016multimodal}     & 0.913 & -          & 0.918 &-          & 0.907 & -          & 0.915 &-            & 0.911 &-                           & 0.914 &-                        \\ \hline
\end{tabular}

\label{tab:chalearn-results}
\end{table*}

Since \emph{RAM} can also predict the apparent personality attributes from the output of \emph{EAM}, as shown in Fig.~\ref{fig:overview}, it can provide our personality network with complementary information. To demonstrate this, we fuse the predicted attributes of both \emph{RAM} and \emph{PAM}; we use late fusion by a weighted average which assigns a weight of 6 to the personality network and 1 to the \emph{RAM} \ZL{(weights are obtained by performing a grid search on a separate validation set)}. The results  are presented in Table~\ref{tab:personality-comp} as ``\emph{PAM}+\emph{RAM}''.  In this case, we observe another performance boost and the highest overall accuracy.

\subsection{Emotion-to-Apparent-Personality Relationship}

Here we show the possibility of determining apparent personality traits from 2-dimensional affective components.  As can be noticed in Table~\ref{tab:personality-comp} under ``Ours (\emph{RAM})'', we achieve satisfactory apparent personality predictions with only 2-dimensional arousal-valence inputs. 

\begin{figure*}[ht!]
	\centering
    \begin{tabular}{cc}
    \includegraphics[width=0.45\textwidth]{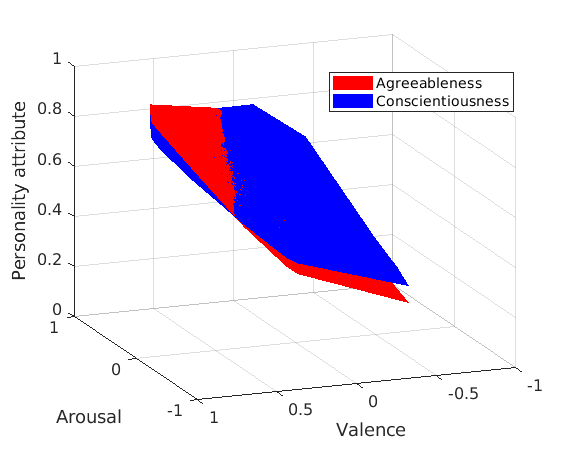}&
	\includegraphics[width=0.45\textwidth]{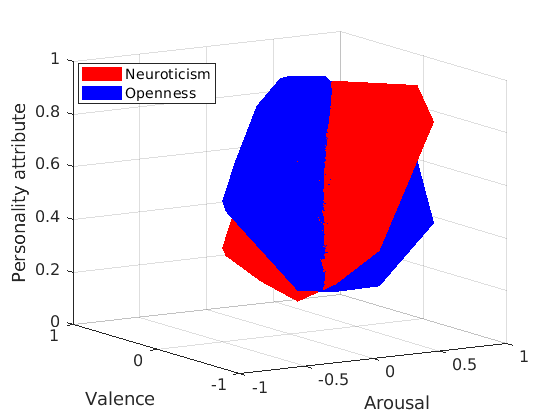}\\
    \end{tabular}
    \caption{Illustration of the relationship between various apparent personality traits and the arousal-valence emotion space, acquired from the input and output of \emph{RAM}. Best viewed in color.}
\label{fig:result-relationship}
\end{figure*}

An illustration of the \relationship is shown in Fig.~\ref{fig:result-relationship}, where each ``disk'' represents a certain apparent personality trait with respect to the corresponding values of arousal and valence. The discoveries are consistent with~\cite{yik2001predicting}: Agreeableness and Conscientiousness are fairly near each other (the two traits share similar emotions), while Neuroticism is located far away from Openness. The ``disk'' for Extraversion (not shown in the Figure) is close to Agreeableness. 
This demonstrates that our $RAM$ network indeed has the ability of learning the \relationship. Based on this, we believe that \proposed can serve as a strong practical baseline for automatically annotating apparent personality based on arousal and valence.
\begin{figure*}[ht!]
\centering
    \begin{tabular}{cc}
    \includegraphics[width=0.44\textwidth]{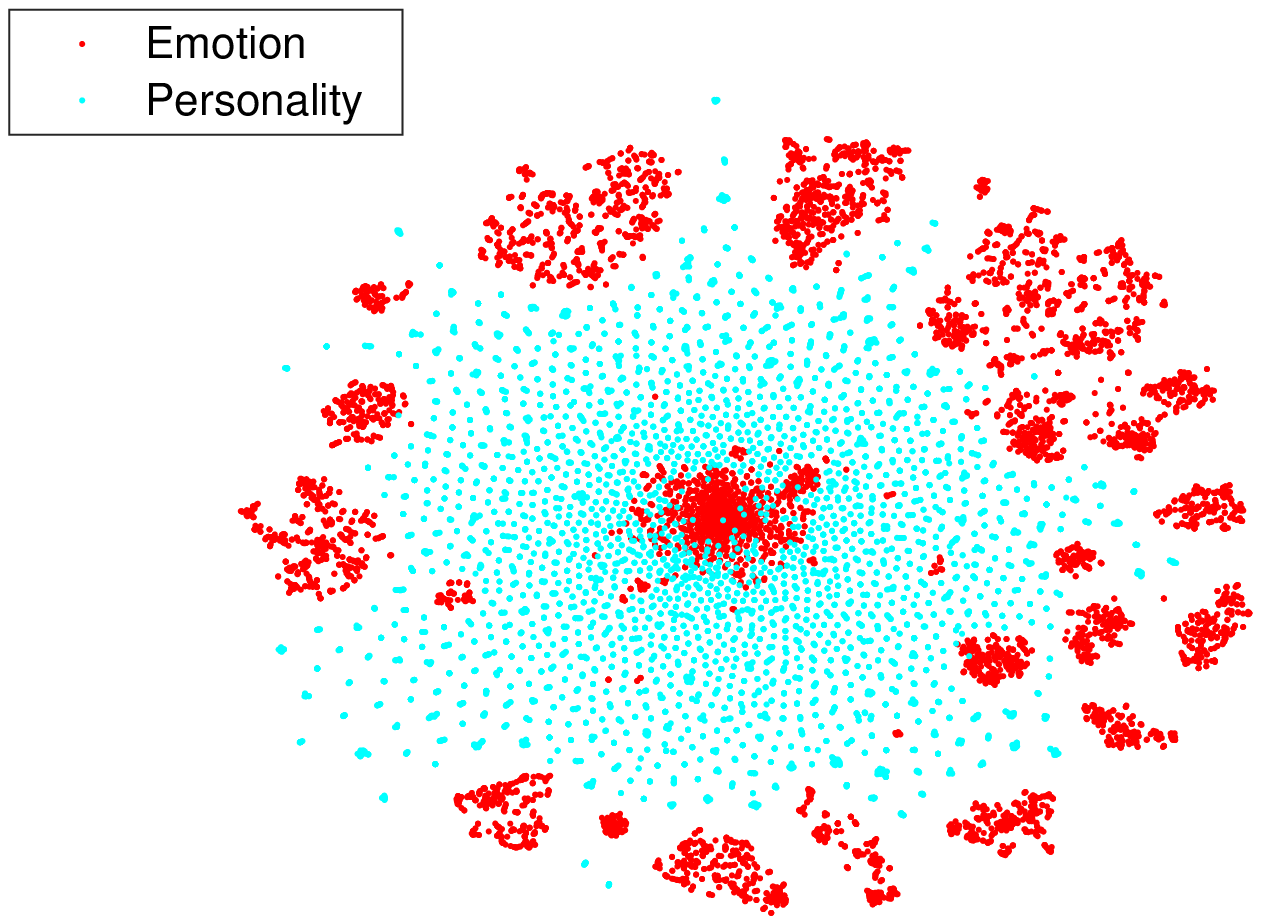}&
	\includegraphics[width=0.44\textwidth]{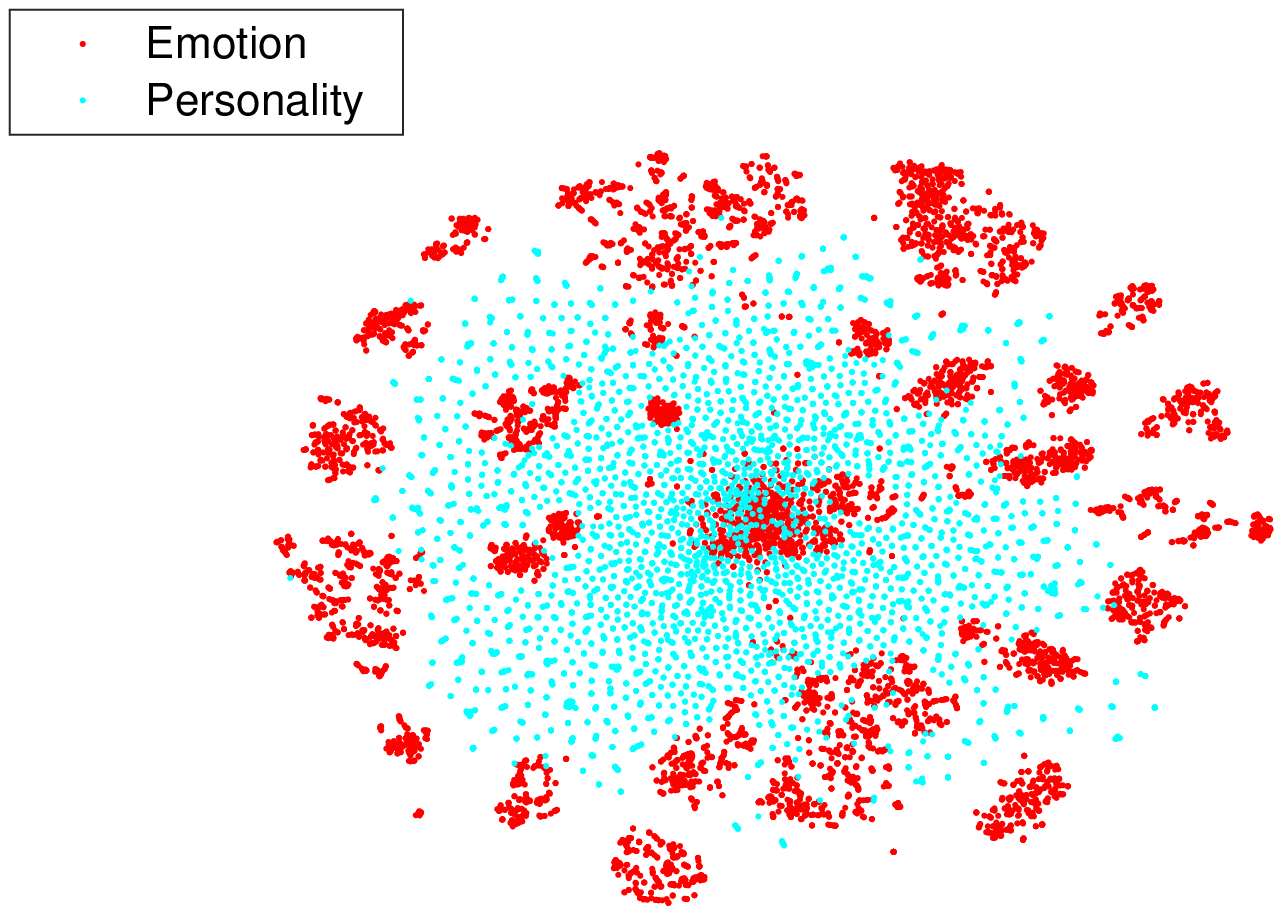}\vspace{-1em}\\
    (a) Without coherence &(b) With coherence\\
    \vspace{-1em}
    \end{tabular}
    \caption{Visualization of the distribution of learned features in~\proposed for both emotion and apparent personality datasets with and without coherence strategy. \ZL{Using the coherence strategy, a large number of features from the emotion dataset are pulled inside the ring, making the two distributions more similar and increasing the overlap between the distributions significantly}. Zoom in for more details. Best viewed in color. }
\label{fig:coherence}
\end{figure*}

\subsection{Ablation Study}

\subsubsection{Effectiveness of Joint Training}
Our novel multi-task learning approach aims to learn a generalizable representation, which is applicable not only to the task in question, but also to other tasks with significant commonalities. In \proposed, since a shared \emph{FEM} is employed by all tasks, additional tasks act as  regularization, which requires the system to perform well on a related task. The backpropagation training from different tasks will  directly impact the representation learning of shared parameters. It prevents overfitting by solving all tasks jointly and allowing for the exploitation of additional training data.  

Table~\ref{tab:jointly} illustrates the effectiveness of this strategy.  As the annotations for the test set of Aff-Wild are not released, we divide the original training set into training and validation set with a ratio of $10:1$ and evaluate all models on the validation set for the emotion task using MSE. \zl{We use the symbols  \cmark \ and \xmark \ to represent  cases where the corresponding functionality is enabled or disabled, respectively.  \zl{ The $2^{nd}$  and $3^{rd}$ row in Table~\ref{tab:jointly} shows the case where we train \emph{EAM} and \emph{PAM} on top of \emph{FEM} individually.  The $4^{th}$ row shows the results of both tasks
when we disable \emph{RAM}. The last row is our final results when all
modules are activated.} When we compare the $3^{rd}$ with the $1^{st}$ row, we observe an improvement in emotion MSE, which indicates the superiority of jointly training emotion with apparent personality. Similarly, an improvement in apparent personality MSE ($3^{rd}$ vs.\ $2^{nd}$ row) verifies that such a strategy is reciprocal. Finally, incorporating the joint training of \emph{RAM} improves the results in both tasks.} We believe these improvements originate from the back-propagation training of CNN, during which the shared parameters within the \emph{FEM} directly impact the generalization ability of the whole system. 


\begin{table}[ht!]
\centering
\caption{Effectiveness of jointly training  \proposed.}
\label{tab:jointly}
\begin{tabular}{ccc|cc}\hline
\multicolumn{3}{c}{Modules in Training} & \multicolumn{2}{c}{MSE in Prediction} \\
\hline
 & Apparent & & & Apparent \\
 Emotion&Personality &Relationship &Emotion  &Personality  \\
 \hline
 \cmark& \xmark & \xmark &0.096  & - \\
 \xmark& \cmark & \xmark & - &0.057 \\
 \cmark& \cmark & \xmark & 0.080&0.033 \\
 \cmark&\cmark&\cmark&0.071&0.027\\\hline
\end{tabular}
\end{table}

\subsubsection{Consensus Function $\mathcal{G}$}
Average temporal pooling has been reported to work well in modeling long-term temporal dependencies for deeply learned representations by \cite{wang2016temporal}. This is also in line with our empirical results on apparent personality recognition. To demonstrate this, we compare average pooling with two other alternatives. One is max pooling, which helps to select the most salient information in its receptive field and has been heavily encoded in popular network structure such as ResNet, VGG and so on. The other is recurrent aggregation, for which we choose the popular LSTM~\citep{hochreiter1997long}. LSTM has been shown to work better than conventional recurrent networks due to its learnable memory gate to avoid gradient vanishing or explosion. 

In our implementation, both feature representations from~\emph{FEM} as well as LSTM are jointly optimized. \zl{More specifically, we train LSTM to aggregate all the feature maps from \emph{FEM} with 10 input frames from the apparent personality dataset. The hidden neurons of LSTM are set to 128. After that, we use average pooling to integrate all temporal information, as done in~\cite{mclaughlin2016recurrent}. Finally, a FC layer is employed to directly regress the apparent personality scores.} We achieve an \ZL{average} accuracy of $91.4\%$, $90.6\%$ and  $90.1\%$ for average pooling, max pooling and LSTM, respectively. Max pooling performs worse than average pooling and better than LSTM. This indicates that selecting the most salient information from a video frame does not necessarily capture its overall statistics better. The reason for the weakness of LSTM could be that apparent personality is an orderless concept where temporal dependencies may not be so relevant. 

\subsubsection{Number of Segments $K$}
In our implementation, $K=10$. We empirically find that the apparent personality results are not sensitive when $K$ is within $[5,20]$. However, when both emotion and personality network are jointly optimized, we observe that a balanced input is always beneficial in both tasks. We use a batch size of $100$ for both emotion and apparent personality datasets. In this way, $10$ input videos for apparent personality are used in each batch.  Setting $K$ to a larger value, for example $100$, will lead to  a lower number of either input videos for apparent personality or emotion frames. This further reduces the final performance in both tasks.

\subsubsection{Coherence Strategy}
As reported by~\cite{tzeng2015simultaneous}, a representation with good transferability should be dataset invariant. 
\zl{To verify this, we also report the MSE scores of two tasks when we remove the coherence strategy from \proposed in Table~\ref{tab:coherence}. } We observe that this strategy leads to about $15\%$ improvement in terms of MSE for emotion (Aff-Wild) and apparent personality (Chalearn). 

We visualize the distribution of the deeply learned features from \emph{FEM} (the \emph{fc}$5$ layer of \emph{SphereFace}) in Fig.~\ref{fig:coherence}. More specifically, we project the 512-dimensional features on both emotion and apparent personality datasets into 2 dimensional space and visualize their distributions using t-SNE~\citep{maaten2008visualizing}. \zl{t-SNE visualizes high-dimensional data by giving each data point a location in a two- or three-dimensional map. The visualizations produced by t-SNE are often significantly better than other advanced techniques}. Without a coherence strategy, distributions of those deep features on different datasets can be well classified, i.e.\ features from emotion dataset are mainly distributed in the outer ring of the $x/y$ plane. Using the coherence strategy, a large number of features from the emotion dataset are pulled inside the ring, making the two distributions more similar \zl{and their overlap significantly larger.}

 \begin{table}[htb!]
\renewcommand{\tabcolsep}{2.5pt}
\centering
\caption{\zl{Effectiveness of coherence strategy in \proposed in terms of MSE.}}
\label{tab:coherence}
\begin{tabular}{rcc}\hline
 MSE &With Coherence&Without Coherence  \\\hline
 Emotion&0.071&0.082 \\
 Apparent Personality& 0.027 & 0.032 \\\hline
\end{tabular}
\end{table}

\section{Conclusions}

For the first time, we investigate the feasibility of jointly analyzing apparent personality, emotion, and their relationship within a single deep neural network. This is challenging due to the scarceness of datasets which encompass both emotion and apparent personality annotations. To tackle this issue we propose \proposed, an end-to-end trainable deep network with two CNN branches called emotion and apparent personality network. With shared bottom feature extraction layers, these two networks regularize each other within a multi-task learning framework, where each one is dedicated to their own annotated dataset. We further employ an adversarial-like loss function  to promote representation coherence between heterogeneous dataset sources, which leads to further performance boosts.  We demonstrate the effectiveness of \proposed on two apparent personality and emotion datasets. We find that the proposed joint training of both emotion and apparent personality networks can lead to a more generalizable representation for both tasks.


\section*{Acknowledgement}
This research is supported by the SERC Strategic Fund from the Science and Engineering Research Council (SERC), A*STAR (project no. a1718g0046).

\bibliographystyle{IEEEtran}
\bibliography{egbib}

\begin{IEEEbiography}[{\includegraphics[width=1in,height=1.25in,clip,keepaspectratio]{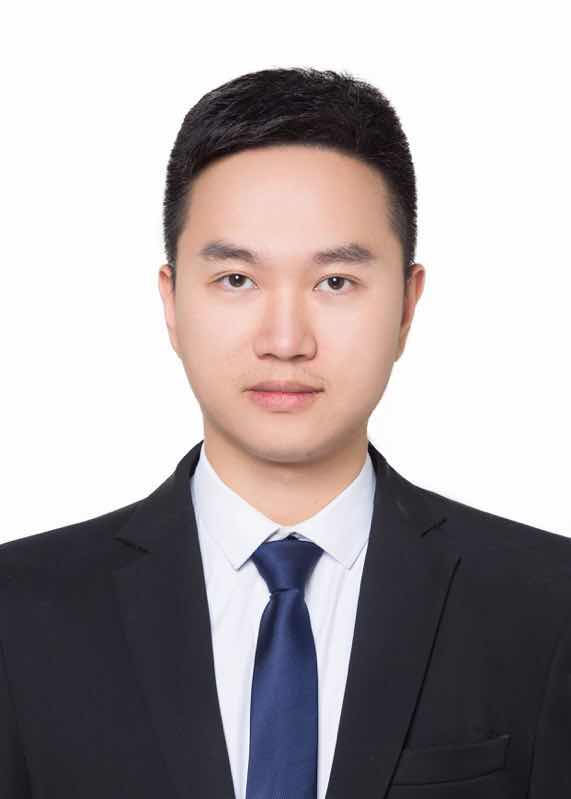}}]{Le Zhang}received the B.Eng.\ degree from the University of Electronic Science and Technology of China (UESTC)
in 2011. He received his M.Sc.\ and Ph.D.\ degrees from Nanyang Technological University (NTU) in
2012 and 2016, respectively. Currently, he is a scientist at Institute for Infocomm Research, Agency
for Science, Technology and Research (A*STAR), Singapore. Prior to that, he was a postdoctoral researcher at the University of Illinois' Advanced Digital Sciences Center (ADSC), Singapore. His current research interests include deep learning and computer vision.
\end{IEEEbiography}

\begin{IEEEbiography}[{\includegraphics[width=1in,height=1.25in,clip,keepaspectratio]{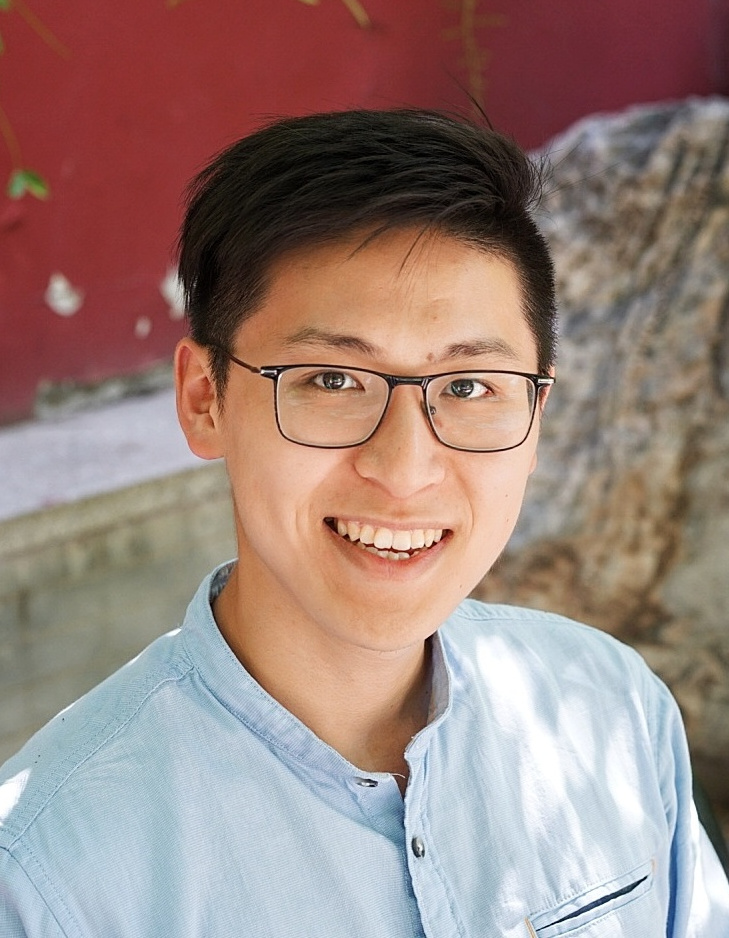}}]{Songyou Peng} is currently a PhD student in Max Planck ETH Center for Learning Systems, a joint program between ETH Zurich and the Max Planck Institute for Intelligent Systems. He received the Erasmus Mundus M.Sc.\ in Computer Vision and Robotics in 2017 and received B.Eng degree from Xi'an Jiaotong University in 2015. He was a research engineer at Institute for Infocomm Research, A*STAR, Singapore. Prior to that, he was a research engineer at Advanced Digital Sciences Center, a research center of the University of Illinois at Urbana-Champaign. His research interests are computer vision and machine learning.
\end{IEEEbiography}

\begin{IEEEbiography}[{\includegraphics[width=1in,height=1.25in,clip,keepaspectratio]{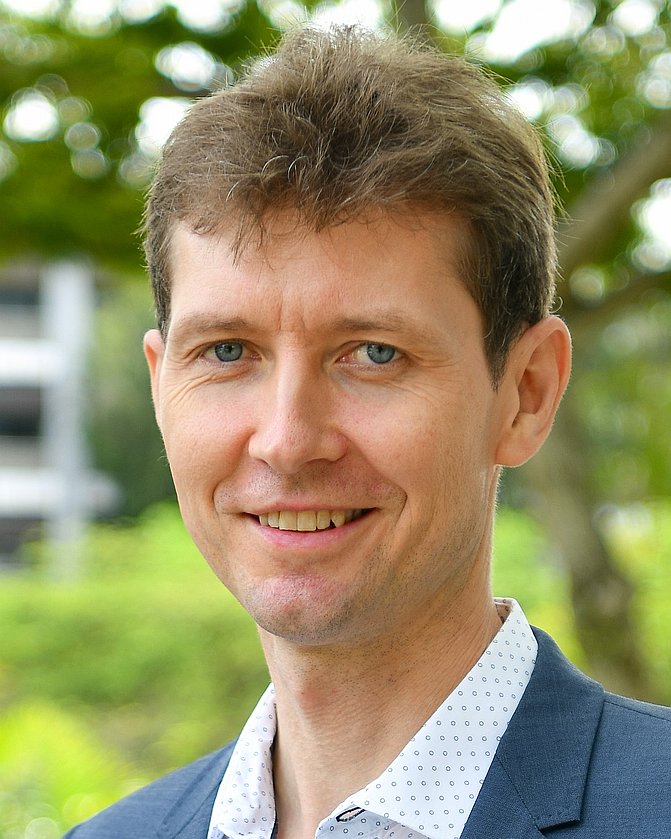}}]{Stefan Winkler} is Deputy Director at AI Singapore and Associate Professor at the National University of Singapore (NUS). Prior to that, he was Distinguished Scientist and Program Director at the University of Illinois' Advanced Digital Sciences Center (ADSC) in Singapore. He also co-founded two start-ups and worked for a Silicon Valley company. He has a Ph.D.\ degree from the Ecole Polytechnique F{\'e}d{\'e}rale de Lausanne (EPFL), Switzerland, and a Dipl.-Ing.\ (M.Eng./B.Eng.) degree from the University of Technology Vienna, Austria. He is an IEEE Fellow and has published over 150 papers. His research interests include video processing, computer vision, machine learning, perception, and human-computer interaction.
\end{IEEEbiography}

\end{document}